\documentclass[letterpaper, 10 pt, conference]{ieeeconf}

\usepackage{amsmath,amsfonts}
\usepackage{algorithmic}
\usepackage{algorithm}
\usepackage{array}
\usepackage[caption=false,font=normalsize,labelfont=sf,textfont=sf]{subfig}
\usepackage{textcomp}
\usepackage{stfloats}
\usepackage{url}
\usepackage{verbatim}
\usepackage{graphicx}
\usepackage{cite}
\usepackage{siunitx}
\usepackage{cleveref}
\usepackage{xcolor}
\usepackage[acronym]{glossaries}
\usepackage{graphicx}
\usepackage{amssymb}
\usepackage{array}
\usepackage{booktabs}
\usepackage{comment}

\usepackage{tikz}
\usepackage{pgfplots}
\usepackage{graphicx}
\usepackage{standalone} %

\pgfplotsset{compat=1.18}
\usepgfplotslibrary{units}
\usetikzlibrary{positioning,intersections, hobby, patterns, calc, decorations.pathmorphing, decorations.markings, shadows, shapes, plotmarks, shapes.multipart}
\usetikzlibrary{fit}
\usetikzlibrary{arrows.meta}

\usepgfplotslibrary{external}

\tikzset{
  external/system call={pdflatex \tikzexternalcheckshellescape --halt-on-error --interaction=batchmode --jobname "\image" "\texsource"}
}
\tikzexternaldisable %

\newenvironment{externalize}[1]{
\tikzexternalenable %
\tikzsetnextfilename{img-#1} %
}{}

\pgfplotsset{
            /pgfplots/layers/niceLayers/.define layer set={
                        axis background,
                        axis grid,
                        main,
                        axis ticks,
                        axis lines,
                        axis tick labels,
                        axis descriptions,
                        axis foreground
            }{/pgfplots/layers/standard}
}

\pgfplotsset{
            every axis/.append style={
                        set layers=niceLayers,
                        tick label style={font=\scriptsize},
                        clip marker paths=true,
                        line width=1pt,
                        line cap=round,
                        line join=round,
                        tick style={semithick, color=black},
                        legend style={
                        		   font=\scriptsize,
                                    /tikz/every even column/.append style={column sep=2mm},
                                    cells={anchor=west}, %
                        },
                        xmajorgrids,
                        ymajorgrids,
            }
}

\pgfkeys{/pgf/number format/.cd,1000 sep={}}

\newacronym[plural=PUs, firstplural=power units]{pu}{PU}{power unit}
\newacronym{f1}{F1}{Formula 1}
\newacronym{cfd}{CFD}{computational fluid dynamics}
\newacronym{mpc}{MPC}{model predictive control}
\newacronym{drs}{DRS}{Drag Reduction System}
\newacronym{mguk}{MGU-K}{motor-generator unit -- kinetic}
\newacronym{mguh}{MGU-H}{motor-generator unit -- heat}
\newacronym{nn}{NN}{neural network}
\newacronym[plural=OCPs, firstplural=optimal control problems]{ocp}{OCP}{optimal control problem}
\newacronym[plural=NLPs, firstplural=nonlinear programs]{nlp}{NLP}{nonlinear program}
\newacronym{kkt}{KKT}{Karush-Kuhn-Tucker}
\newacronym{mpcc}{MPCC}{mathematical program with complementarity constraints}
\newacronym{licq}{LICQ}{linear independence constraint qualification}
\newacronym{mfcq}{MFCQ}{Mangasarian-Fromovitz constraints qualification}
\newacronym{svo}{SVO}{Social Value Orientation}
\newacronym{ibr}{IBR}{iterated best-response}
\newacronym{mltp}{MLTP}{minimum lap time problem}
\newacronym{qss}{QSS}{quasi-steady-state}
\newacronym{em}{EM}{energy management}
\newacronym[plural=GNEPs, firstplural=Generalized Nash Equilibrium Problems]{gnep}{GNEP}{Generalized Nash Equilibrium Problem}
\newacronym[plural=HEVs, firstplural=hybrid-electric vehicles]{hev}{HEV}{hybrid-electric vehicle}
\newacronym{soc}{SOC}{state-of-charge}
\newacronym{ice}{ICE}{internal combustion engine}
\newacronym{ers}{ERS}{engine recovery system}
\newacronym{rl}{RL}{reinforcement learning}
\newacronym{minlp}{MINLP}{mixed-integer nonlinear program}
\newacronym{mdp}{MDP}{Markov decision process}
\newacronym{sac}{SAC}{soft actor-critic}
\newacronym{des}{DES}{discrete-event simulation}
\newacronym{dp}{DP}{dynamic programming}
\newacronym{sa}{SA}{single-agent}

\newcommand{\isMainDocument}{}
\newcommand{\DeltaEb}{\Delta E_\mathrm{b,all}}
\newcommand{\DeltaEf}{\Delta E_\mathrm{f,all}}
\newcommand{\TW}{\mathrm{TW}}
\newcommand{\TC}{\mathrm{TC}}
\newcommand{\TA}{\mathrm{TA}}
\newcommand{\PS}{\mathrm{PS}}
\newcommand{\bComp}{b_\mathrm{comp}}

\newlength{\myfigskip}
\begin{document}

\title{\LARGE \bf Learning-based Multi-agent Race Strategies in Formula 1}
\author{Giona Fieni$^{1}$, Joschua W\"uthrich$^{1}$, Marc-Philippe Neumann$^{1}$, Christopher H. Onder$^{1}$
\thanks{$^{1}$Institute for Dynamic Systems and Control, ETH Z\"urich, Z\"urich, Switzerland,
{\tt\small gfieni@ethz.ch}}}

\maketitle

\begin{abstract}
In Formula 1, race strategies are adapted according to evolving race conditions and competitors’ actions. This paper proposes a reinforcement learning approach for multi-agent race strategy optimization. Agents learn to balance energy management, tire degradation, aerodynamic interaction, and pit-stop decisions. Building on a pre-trained single-agent policy, we introduce an interaction module that accounts for the behavior of competitors. The combination of the interaction module and a custom self-play training scheme generates competitive policies, and agents are ranked based on their relative performance. Results show that the agents adapt pit timing, tire selection, and energy allocation in response to opponents, achieving robust and consistent race performance. Because the framework relies only on information available during real races, it can support race strategists' decisions before and during races.
\end{abstract}

\section{Introduction}\label{sec:intro}
\gls{f1} is the most famous motorsport. Each year, 22 drivers from 11 teams compete in over 20 races. Teams optimize every aspect of performance, from car development and \gls{pu} operation to strategic planning before and during races. Drivers, in turn, must consistently extract maximum performance while minimizing errors. However, race conditions often deviate from predictions, making rapid decision-making under uncertainty essential. In such situations, the experience of race engineers is extremely important.

Since 2014, \gls{f1} has adopted hybrid-electric \glspl{pu}. The \gls{ice} and the \gls{mguk} operate in synergy to generate the propulsive power. In addition to fuel management, the electrical energy stored in the battery must be carefully deployed. As fuel mass directly affects vehicle performance -- lighter cars are faster -- energy allocation becomes a critical performance factor.

Race strategies mostly influence the race outcome. Regulations require the use of at least two different tire compounds during a race. When tire performance deteriorates, the pit wall calls for a \textit{pit stop}. The performance gain of new tires must compensate the time lost in the pit lane.

Despite extensive simulations, it is impossible to predict every scenario. Teams typically compute a large number of Monte Carlo simulations to prepare three to four baseline strategies, which are continuously adapted according to the evolving race context. This includes observing opponents' performance and react to their strategic decisions.  

In this paper, our goal is to support the online decision-making process with algorithms. In \gls{f1}, decisions have to be taken within seconds: Rather than predicting, the focus is to robustly react to unforeseen events while accounting for the active response of a competitor. 

\subsection{Related work}\label{subsec:literaturereview}
For the research literature related to this paper, we consider \gls{sa} race strategies and its multi-agent perspective. 

The first part can be subdivided in simulations \cite{heilmeier2020virtual}, optimizations \cite{duhr2023minimum,van2022maximum,bonomi2023evolutionary,heine2023optimization,neumann2024strategic}, and learning-based methods \cite{boettinger2023mastering,thomas2025explainable,liu2021formula,fieni2025towards}. 
In \cite{heilmeier2020virtual}, \glspl{nn} used as virtual strategy engineer in race simulations delivered results close to reality. Optimizations deal with energy management \cite{duhr2023minimum}, charging stops in endurance racing \cite{van2022maximum}, evolutionary algorithms \cite{bonomi2023evolutionary} and co-design \cite{neumann2024strategic} for pit stops. Stochastic scenarios optimized via \gls{dp} are studied in \cite{heine2023optimization}. Learning-based methods mainly employ \gls{rl} to tackle the problem. In particular, \cite{fieni2025towards} considers both energy management and pit stops. 

When considering competitors' interaction, it has to be distinguished between competitors' awareness and multi-agent methods. Building on \cite{paparusso2022competitors}, a \gls{mpc} framework for competitor-aware race strategies is developed in \cite{van2024model}, and a two-player game is solved in \cite{aguad2024optimizing} with \gls{dp}. While pit stops are optimized, the energy management is neglected. Multi-agent \gls{rl} in Formula~E racing is explored by \cite{liu2024formula}. Following an earlier version of the present work, the subsequent study \cite{de2026competitor} adapted our setup for electric endurance racing. However, different regulations and requirements render these problems substantially different: The racing model and the state transitions are heavily simplified, pit stops are not driven by tire degradation and the performance of different tire compounds is neglected. 
 
We aim to bridge the gap towards multi-agent race strategies accounting for active competitors' response. In particular, the framework incorporates energy management, pit stops, and tire degradation. \gls{f1} represents a general case of race strategies problems, which are also present, in simplified form, in other racing series. Given the non-smooth and multi-agent system, classical optimization methods encounter practical limitations, whereas \gls{rl} emerges as promising tool. Our approach builds upon \cite{fieni2025towards}, where the nominal policy is benchmarked against an optimization framework. 
 
\subsection{Contributions}\label{subsec:contributions}
We develop a framework for the training and simulation of multi-agent race strategies, by bridging the gap between the \gls{sa} \gls{rl} approach of \cite{fieni2025towards} and a game with competitor-reacting agents. In particular, we contribute as follows. 

First, we extend the environment of \cite{fieni2025towards} including a second agent and a model describing their physical coupling.

Second, by augmenting the policy with an interaction module and employing a custom self-play training scheme, we generate agents able to account for the active response of a competitor. We observe the agents exploiting typical \gls{f1} strategies such as the \textit{undercut}.

Third, we show the superiority of the interaction module compared to the baseline \gls{sa} policy, and that generated agents adapt their race strategy according to the opponents, in terms of energy management, pit stops and tire compound selection.  

Because the agents rely only on information available during real races, their outputs can support race strategist to improve their decision-making process. 

\subsection{Outline}\label{subsec:structure}
This paper is organized as follows: \Cref{sec:RLsetup} introduces the \gls{rl} environment, the interaction model and the agent structure that accounts for interactions. \Cref{sec:MAextension} presents the extension to the multi-agent framework, including the training scheme and the generation of high-performing agents. 
Finally, in \Cref{sec:results} we let the agents compete with each other and we conclude the paper in \Cref{sec:conclusions}.

\section{Reinforcement learning setup}\label{sec:RLsetup}
In this section, we describe the setup for the \gls{rl} framework. First, we present the environment and its submodules. Then, we motivate the choice for the agent's architecture and formulate the \gls{mdp}. 

\subsection{Environment}\label{subsec:RLenv}
Building upon \cite{fieni2025towards}, we extend the framework to handle multi-agent interactions. One episode corresponds to a race, which is discretized on a lap-by-lap basis. We denote with $k\in[0,\dots,N_\mathrm{laps}]$ the lap number. 

\Cref{fig:environment} shows a schematic of the environment and the information exchange. Agent $1$ interacts with the environment using the action vector $\mathbf{a}^{1}$, and it receives the observations $\mathbf{o}^{1}$ and $\mathbf{\tilde o}^{i}$. To include the active response of agent $i$ embedded in the environment, actions are sampled from its policy based on its observations. This framework models a dynamic game, in which agents observe and actively account for the closed-loop response of the opponents. 
\begin{figure*}
	\centering
	\scalebox{0.95}{
	\begin{externalize}{Environment}
		\begin{tikzpicture}[scale=1, every node/.append style={outer sep=0pt}, >=stealth, font=\small]

\ifdefined\isMainDocument

\else
	\newcommand{\DeltaEb}{\Delta E_\mathrm{b,all}}
	\newcommand{\DeltaEf}{\Delta E_\mathrm{f,all}}
	\newcommand{\TW}{\mathrm{TW}}
	\newcommand{\TC}{\mathrm{TC}}
	\newcommand{\PS}{\mathrm{PS}}
	\newcommand{\bComp}{b_\mathrm{compound}}
\fi

\tikzstyle{block} = [draw, rectangle, minimum height=1cm, minimum width=1cm]
\def\dCirc{0.1cm}
\def\radius{0.2cm}
\tikzstyle{dot2} = [draw, circle, minimum size=\dCirc, fill = black, inner sep=0cm]
\tikzstyle{dot3} = [draw, circle, minimum size=\radius, color=gray!20, fill = gray!20, inner sep=0cm]

\def\altezza{1cm}
\def\altezzatwo{1.3cm}
\def\altezzathree{1.8cm}
\def\larghezza{2.8cm}
\def\vertDist{0.5cm}
\def\vertDisttwo{0.65cm}
\def\horzDist{3.8cm}
\def\shadingOffset{0.06cm}
\def\textOffset{0.2cm}
\def\heightRef{0.07cm}
\def\widthRef{0.3cm}

\def\larghezzaML{0.8*\larghezza}
\def\larghezzaAgent{0.65*\larghezza}
\def\vertOffset{-\vertDist + \altezza}
\def\vertOffsetSecond{-2*\vertDist + \altezza}
\def\vertOffsetThird{-4*\vertDist + \altezza}
\def\oneThirdUp{\vertDist}
\def\twoThirdUp{2*\vertDist/3}

\node [coordinate] at (0,0) (agent1) {};
\node [block, fill=red!10, minimum height=\altezza, minimum width=\larghezzaAgent] at (agent1) (Agent1Block) {Agent 1};

\node [coordinate] at ($(Agent1Block) + (5*\larghezzaAgent, 0)$) (Env) {};
\node [block, fill=gray!20, draw=none, minimum height=4.2*\altezza, minimum width=6.5*\larghezzaML] at (Env) (EnvBlock) {};
\node at ($(Env) + (0, 1.8*\altezza)$) (TextEnv) {\textbf{Environment}};

\node [coordinate] at ($(Agent1Block) + (1.7*\larghezzaAgent, 0)$) (SAMR) {};
\node [block, fill=black, minimum height=\altezzatwo, minimum width=\larghezzaML] at ($(SAMR) + (\shadingOffset,-\shadingOffset)$) () {};
\node [block, fill=green!10, minimum height=\altezzatwo, minimum width=\larghezzaML,align=center] at (SAMR) (SAMRBlock) {Race car 1};

\node [coordinate] at ($(SAMRBlock) + (2.5*\larghezzaAgent, 0)$) (AIM) {};
\node [block, fill=black, minimum height=\altezzathree, minimum width=\larghezzaML] at ($(AIM) + (\shadingOffset,-\shadingOffset)$) () {};
\node [block, fill=white, minimum height=\altezzathree, minimum width=\larghezzaML, align=center] at (AIM) (AIMBlock) {Aerodynamic\\interaction};

\node [coordinate] at ($(AIMBlock) + (2.5*\larghezzaAgent, 0)$) (SAMR2) {};
\node [block, fill=black, minimum height=\altezzatwo, minimum width=\larghezzaML] at ($(SAMR2) + (\shadingOffset,-\shadingOffset)$) () {};
\node [block, fill=green!10, minimum height=\altezzatwo, minimum width=\larghezzaML, align=center] at (SAMR2) (SAMR2Block) {Race car $i$};

\node [coordinate] at ($(SAMR2Block) + (1.7*\larghezzaAgent, 0)$) (agenti) {};
\node [block, fill=red!10, minimum height=\altezza, minimum width=\larghezzaAgent] at (agenti) (AgentiBlock) {Agent $i$};

\node [coordinate] at ($(Agent1Block.north) + (0, \vertDisttwo)$) (Agent1Sopra) {};
\node [coordinate] at ($(Agent1Block.south) + (-\larghezzaAgent/6, -\vertDisttwo-\oneThirdUp/2)$) (Agent1SottoSinistra) {};
\node [coordinate] at ($(Agent1Block.south) + (\larghezzaAgent/6, -\vertDisttwo)$) (Agent1SottoDestra) {};

\node [coordinate] at ($(SAMRBlock.north) + (0, \vertDist)$) (SAMRSopra) {};
\node [coordinate] at ($(SAMRBlock.south) + (-\larghezzaML/6, -\vertDist)$) (SAMRSottoSinistra) {};
\node [coordinate] at ($(SAMRBlock.south) + (\larghezzaML/6, -\vertDist-\oneThirdUp)$) (SAMRSottoDestra) {};
\node [coordinate] at ($(SAMRBlock.east) + (0, \altezzatwo/6)$) (SAMRDestraSopra) {};
\node [coordinate] at ($(SAMRBlock.east) + (\shadingOffset, -\altezzatwo/6)$) (SAMRDestraSotto) {};

\node [coordinate] at ($(AIMBlock.west) + (0, \altezzatwo/6)$) (AIMSinistraSopra) {};
\node [coordinate] at ($(AIMBlock.west) + (0, -\altezzatwo/6)$) (AIMSinistraSotto) {};
\node [coordinate] at ($(AIMBlock.east) + (\shadingOffset, \altezzatwo/6)$) (AIMDestraSopra) {};
\node [coordinate] at ($(AIMBlock.east) + (0, -\altezzatwo/6)$) (AIMDestraSotto) {};

\node [coordinate] at ($(SAMR2Block.north) + (0, \vertDist)$) (SAMR2Sopra) {};
\node [coordinate] at ($(SAMR2Block.south) + (-\larghezzaML/6, -\vertDist-\oneThirdUp/2)$) (SAMR2SottoSinistra) {};
\node [coordinate] at ($(SAMR2Block.south) + (\larghezzaML/6, -\vertDist)$) (SAMR2SottoDestra) {};
\node [coordinate] at ($(SAMR2Block.west) + (0, \altezzatwo/6)$) (SAMR2SinistraSopra) {};
\node [coordinate] at ($(SAMR2Block.west) + (0, -\altezzatwo/6)$) (SAMR2SinistraSotto) {};

\node [coordinate] at ($(AgentiBlock.north) + (0, \vertDisttwo)$) (AgentiSopra) {};
\node [coordinate] at ($(AgentiBlock.south) + (-\larghezzaAgent/6, -\vertDisttwo)$) (AgentiSottoSinistra) {};
\node [coordinate] at ($(AgentiBlock.south) + (\larghezzaAgent/6, -\vertDisttwo-\oneThirdUp)$) (AgentiSottoDestra) {};

\node [coordinate] at ($(SAMRSottoDestra) + (0, \oneThirdUp/2)$) (carmassSopraIntersect) {};

\draw (Agent1Block.north) to (Agent1Sopra);
\draw (Agent1Sopra) to node[midway, above] {$\mathbf{a}^1$} (SAMRSopra);
\draw[->] (SAMRSopra) to (SAMRBlock.north);

\draw ($(SAMRBlock.south) + (-\larghezzaML/6, 0)$) to (SAMRSottoSinistra);
\draw (Agent1SottoDestra) to node[midway, above] {$\mathbf{o}^1$} (SAMRSottoSinistra);
\draw[->] (Agent1SottoDestra) to ($(Agent1Block.south) + (\larghezzaAgent/6, 0)$);

\draw ($(SAMR2Block.south) + (-\larghezzaML/6, 0)$) to (SAMR2SottoSinistra);
\draw (Agent1SottoSinistra) to node[midway, above] {$\mathbf{\tilde o}^i$} (SAMR2SottoSinistra);
\draw[->] (Agent1SottoSinistra) to ($(Agent1Block.south) + (-\larghezzaAgent/6, 0)$);
\node[dot3] at (carmassSopraIntersect) {};

\draw ($(SAMRBlock.south) + (\larghezzaML/6, 0)$) to (SAMRSottoDestra);
\draw (SAMRSottoDestra) to node[midway, below] {$\mathbf{\tilde o}^1$} (AgentiSottoDestra);
\draw[->] (AgentiSottoDestra) to ($(AgentiBlock.south) + (\larghezzaAgent/6, 0)$);

\draw ($(SAMR2Block.south) + (\larghezzaML/6, 0)$) to (SAMR2SottoDestra);
\draw (SAMR2SottoDestra) to node[midway, above] {$\mathbf{o}^i$} (AgentiSottoSinistra);
\draw[->] (AgentiSottoSinistra) to ($(AgentiBlock.south) + (-\larghezzaAgent/6, 0)$);

\draw (AgentiBlock.north) to (AgentiSopra);
\draw (SAMR2Sopra) to node[midway, above] {$\mathbf{a}^i$} (AgentiSopra);
\draw[->] (SAMR2Sopra) to (SAMR2Block.north);

\draw[->] (SAMRDestraSopra) to node[midway, above] {$T_\mathrm{lap}^1$} (AIMSinistraSopra);
\draw[->] (AIMSinistraSotto) to node[midway, below] {$t_\mathrm{gap}^1$, $\Delta T_\mathrm{int}^1$} (SAMRDestraSotto);

\draw[->] (SAMR2SinistraSopra) to node[midway, above] {$T_\mathrm{lap}^i$}(AIMDestraSopra);
\draw[->] (AIMDestraSotto) to node[midway, below] {$t_\mathrm{gap}^i$, $\Delta T_\mathrm{int}^i$}(SAMR2SinistraSotto);

\end{tikzpicture}
	\end{externalize}
	}
	\caption{Schematic of the agent-environment interaction. Agent $1$ is the agent to be trained, while agent $i$ is fixed and it is part of the environment. With their actions, they directly affect the ego car. Aerodynamic interaction couples the models of the two cars. The observations are divided whether they come from the ego car or from the competitor's one. The original setup of \cite{fieni2025towards} considers only the modules \textit{Agent 1} and \textit{Race car 1}.}
	\label{fig:environment}
\end{figure*}

The race strategy is the collection of pit wall's decisions taken for the car. These are the allocated fuel and battery energy per lap, respectively $\Delta E_\mathrm{f,all}$ and $\Delta E_\mathrm{b,all}$, as well as the pit stop and tire compound decision, which are captured by the decision variable $\PS\in\{0,1,2,3\}$, defined in \Cref{tab:PS}.
\begin{table}
\begin{center}
\caption{Actions corresponding to the $\PS$ variable.}\label{tab:PS}
\begin{tabular}{c c c c c}
\toprule
 $\PS$ & 0 & 1 & 2 & 3\\
\midrule
 \textbf{Action} & do not pit & pit for soft & pit for medium & pit for hard\\
 \bottomrule
\end{tabular}
\end{center}
\end{table}

In our setup, the agent corresponds to the pit wall. For this reason, the action vector is 
\begin{equation} 
\mathbf{a} = \begin{pmatrix}\Delta E_\mathrm{f,all} & \Delta E_\mathrm{b,all} & \PS\end{pmatrix}, 
\end{equation}
which influences directly only the states of the ego car. The dynamics of the single cars evolve accordingly to the blocks \textit{Race car}, presented in \Cref{subsec:SARM}. 

An agent cannot fully observe the states of the opponent. To potentially deploy the agent in real races, only information available to the pit wall of the team can be exploited. Hence, we distinguish between the internal states of the ego car $\mathbf{o}$ and the observable information about the competitor $\mathbf{\tilde o}$, both defined in \Cref{subsec:SARM}. 

Racing in the wake of another car affects the strategy. The inclusion of an interaction model allows to investigate alternative strategies that explicitly account for the presence of another agent. We capture slipstream effects with the block \textit{Aerodynamic interaction}, presented in \Cref{subsec:aeroint}.

\subsection{Race car model}\label{subsec:SARM}
This model describes the evolution of the car's internal states according to the action given by the pit wall, as shown in \Cref{fig:racemodel}. Only for this module, we adopt the same model equations of \cite{fieni2025towards} and consider the following states: the battery energy content $E_\mathrm{b}$, the available fuel energy $E_\mathrm{f}$, the car mass changing due to fuel consumption $m_\mathrm{car}$, the tire compound $\TC$, the tire wear $\TW$ and the total race time $T_\mathrm{race}$. For the sake of space, the model equations are not presented in this paper, but all the calculations happen in the blocks \textit{Vehicle's states} and \textit{Race time}.
\begin{figure}
	\centering
	\scalebox{0.63}{
	\begin{externalize}{RaceModel}
		\begin{tikzpicture}[scale=1, every node/.append style={outer sep=0pt}, >=stealth, font=\small]

\ifdefined\isMainDocument

\else
	\newcommand{\DeltaEb}{\Delta E_\mathrm{b,all}}
	\newcommand{\DeltaEf}{\Delta E_\mathrm{f,all}}
	\newcommand{\TW}{\mathrm{TW}}
	\newcommand{\TC}{\mathrm{TC}}
	\newcommand{\PS}{\mathrm{PS}}
	\newcommand{\bComp}{b_\mathrm{compound}}
\fi

\tikzstyle{block} = [draw, rectangle, minimum height=1cm, minimum width=1cm]
\def\dCirc{0.1cm}
\def\radius{0.2cm}
\tikzstyle{dot2} = [draw, circle, minimum size=\dCirc, fill = black, inner sep=0cm]
\tikzstyle{dot3} = [draw, circle, minimum size=\radius, color=green!10, fill=green!10, inner sep=0cm]

\def\altezza{1cm}
\def\larghezza{2.8cm}
\def\vertDist{2cm}
\def\horzDist{3.8cm}
\def\shadingOffset{0.06cm}
\def\textOffset{0.2cm}
\def\heightRef{0.07cm}
\def\widthRef{0.3cm}

\def\larghezzaTC{1.4*\larghezza}
\def\larghezzaTW{2.3*\larghezza}
\def\larghezzaLaptime{4.14*\larghezza}
\def\vertOffset{-\vertDist + \altezza}
\def\vertOffsetSecond{-2*\vertDist + \altezza}
\def\vertOffsetThird{-4*\vertDist + \altezza}
\def\oneThirdUp{\vertDist/3}
\def\twoThirdUp{2*\vertDist/3}

\node [coordinate] at (0.3cm,0.5cm) (SARM) {};
\node [block, fill=black, minimum height=8.5*\altezza, minimum width=1.10*\larghezzaLaptime] at ($(SARM) + (\shadingOffset,-\shadingOffset)$) () {};
\node [block, fill=green!10, minimum height=8.5*\altezza, minimum width=1.10*\larghezzaLaptime] at (SARM) (SARMBlock) {};
\node at ($(SARM) + (-0.45*\larghezzaLaptime, -3.9*\altezza)$) (TextSARM) {\textbf{Race car}};

\node [coordinate] at (0,0) (laptime) {};
\node [block, fill=white, minimum height=\altezza, minimum width=\larghezzaLaptime] at (laptime) (laptimeBlock) {Lap time};

\node [coordinate] at ($(laptimeBlock) + (0.1111*\larghezzaLaptime, \vertDist)$) (vehicle) {};
\node [block, fill=black, minimum height=\altezza, minimum width=0.8*\larghezzaTW] at ($(vehicle) + (\shadingOffset,-\shadingOffset)$) () {};
\node [block, fill=white, minimum height=\altezza, minimum width=0.8*\larghezzaTW] at (vehicle) (vehicleBlock) {Vehicle's states};

\node [coordinate] at ($(laptime) + (0, -\vertDist)$) (racetime) {};
\node [block, fill=black, minimum height=\altezza, minimum width=\larghezza] at ($(racetime) + (\shadingOffset,-\shadingOffset)$) () {};
\node [block, fill=white, minimum height=\altezza, minimum width=\larghezza] at (racetime) (racetimeBlock) {Race time};

\node [coordinate] at (vehicleBlock.north) (vsi2) {};
\node [coordinate] at ($(vehicleBlock.north) + (-0.5*\larghezzaTW/2,0)$) (vsi1) {};
\node [coordinate] at ($(vehicleBlock.north) + (0.5*\larghezzaTW/2,0)$) (vsi3) {};
\node [coordinate] at ($(vehicleBlock.south) + (-1.5*\larghezzaTW/5,0)$) (vso1) {};
\node [coordinate] at ($(vehicleBlock.south) + (-0.5*\larghezzaTW/5,0)$) (vso2) {};
\node [coordinate] at ($(vehicleBlock.south) + (0.5*\larghezzaTW/5,0)$) (vso3) {};
\node [coordinate] at ($(vehicleBlock.south) + (1.5*\larghezzaTW/5,0)$) (vso4) {};

\node [coordinate] at ($(laptimeBlock.north west) + (\larghezzaLaptime/9,0)$) (lti1) {};
\node [coordinate] at ($(laptimeBlock.north west) + (2*\larghezzaLaptime/9,0)$) (lti2) {};
\node [coordinate] at ($(laptimeBlock.north west) + (3*\larghezzaLaptime/9,0)$) (lti3) {};
\node [coordinate] at ($(laptimeBlock.north west) + (4*\larghezzaLaptime/9,0)$) (lti4) {};
\node [coordinate] at ($(laptimeBlock.north west) + (5*\larghezzaLaptime/9,0)$) (lti5) {};
\node [coordinate] at ($(laptimeBlock.north west) + (6*\larghezzaLaptime/9,0)$) (lti6) {};
\node [coordinate] at ($(laptimeBlock.north west) + (7*\larghezzaLaptime/9,0)$) (lti7) {};
\node [coordinate] at ($(laptimeBlock.north west) + (8*\larghezzaLaptime/9,0)$) (lti8) {};
\node [coordinate] at (laptimeBlock.south) (lto1) {};

\node [coordinate] at (racetimeBlock.north) (rti1) {};
\node [coordinate] at (racetimeBlock.south) (rto1) {};

\node [coordinate] at ($(SARMBlock.north) + (-\larghezzaLaptime/2+2*\larghezzaLaptime/9-0.3cm,0)$) (a) {};
\node [coordinate] at ($(SARMBlock.north) + (0.46*\larghezzaLaptime-0.3cm,0)$) (t) {};

\node [coordinate] at ($(SARMBlock.south) + (-\larghezzaLaptime/3,0)$) (o1) {};
\node [coordinate] at ($(SARMBlock.south) + (\larghezzaLaptime/3,0)$) (o2) {};

\node [coordinate] at ($(vehicleBlock) + (0, \vertDist+\altezza/2 + \twoThirdUp)$) (tirewearSopra) {};
\node [coordinate] at ($(vehicleBlock) + (0, \vertDist+\altezza/2 + \oneThirdUp)$) (tirewearSopraIntersect) {};
\node [coordinate] at ($(vehicleBlock) + (-\larghezzaTW/2 - \larghezza,0)$) (tirewearSopraSinistra) {};
\node [coordinate] at ($(vehicleBlock.south)$) (tirewearAtLapTime) {};

\node [coordinate] at ($(a) + (0,-\vertDist/10)$) (belowa) {};
\node [coordinate] at ($(belowa) + (-\larghezzaLaptime/9,0)$) (leftofbelowa) {};
\node [coordinate] at ($(belowa) + (\larghezzaLaptime/9,0)$) (rightofbelowa) {};

\node [coordinate] at ($(vsi1) + (0,\vertDist/5)$) (abovevsi1) {};
\node [coordinate] at ($(vsi2) + (0,2*\vertDist/5)$) (abovevsi2) {};
\node [coordinate] at ($(vsi3) + (0,3*\vertDist/5)$) (abovevsi3) {};

\node [coordinate] at ($(leftofbelowa |- abovevsi3)$) (leftintersection) {};
\node [coordinate] at ($(belowa |- abovevsi2)$) (middleintersection) {};
\node [coordinate] at ($(rightofbelowa |- abovevsi1)$) (rightintersection) {};

\node [coordinate] at ($(t) + (0,-\vertDist/10)$) (belowt) {};
\node [coordinate] at ($(lti8 |- belowt)$) (ttolti8) {};
\node [coordinate] at ($(belowt) + (\vertDist/2.5,0)$) (ttooutput) {};

\node [coordinate] at ($(rto1) + (0,-\vertDist/2.2)$) (rtotooutput) {};

\node [coordinate] at ($(ttooutput |- rtotooutput)$) (toutput) {};

\draw[->] ($(a)+(0,\vertDist/3)$) to node[midway, left] {$\mathbf{a}$} (a);
\draw[->] ($(t)+(0,\vertDist/3)$) to node[midway, left] {$\Delta T_\mathrm{int}, t_\mathrm{gap}$} (t);

\draw (leftintersection) to (abovevsi3);
\draw (middleintersection) to (abovevsi2);
\draw (rightintersection) to (abovevsi1);
\draw[->] (abovevsi1) to (vsi1);
\draw[->] (abovevsi2) to (vsi2);
\draw[->] (abovevsi3) to (vsi3);

\node[dot3] at ($(middleintersection) + (0,\vertDist/5)$) {}; 
\node[dot3] at ($(rightintersection) + (0,\vertDist/5)$) {}; 
\node[dot3] at ($(rightintersection) + (0,2*\vertDist/5)$) {}; 

\draw (a) to (belowa);
\node[dot2] at (belowa) {}; 
\draw (leftofbelowa) to (rightofbelowa);
\draw[->] (leftofbelowa) to node[midway, left] {$\Delta E_\mathrm{b,all}$} (lti1);
\draw[->] (belowa) to node[midway, left] {$\Delta E_\mathrm{f,all}$} (lti2);
\draw[->] (rightofbelowa) to node[midway, left] {PS} (lti3);

\draw[->] (vso1) to node[midway, left] {$E_\mathrm{b}$} (lti4) {};
\draw[->] (vso2) to node[midway, left] {$m_\mathrm{car}$} (lti5) {};
\draw[->] (vso3) to node[midway, left] {TW} (lti6) {};
\draw[->] (vso4) to node[midway, left] {TC} (lti7) {};

\draw (t) to (belowt);
\node[dot2] at (belowt) {}; 
\draw (belowt) to (ttolti8);
\draw [->] (ttolti8) to node[midway, right] {$\Delta T_\mathrm{lap}$} (lti8);

\draw (belowt) to (ttooutput);
\draw [->] (ttooutput) to node[midway, left, yshift=-2cm] {$t_\mathrm{gap}$} (toutput);

\draw [->] (lto1) to node[midway, left] {$T_\mathrm{lap}$} (rti1);

\draw [->] (rto1) to node[midway, left] {$T_\mathrm{race}$} (rtotooutput);

\draw[->] (o1) to node[midway, left] {$\mathbf{o}$} ($(o1)+(0,-\vertDist/3)$);
\draw[->] (o2) to node[midway, left] {$\mathbf{\tilde o}$} ($(o2)+(0,-\vertDist/3)$);

\end{tikzpicture}
	\end{externalize}
	}
	\caption{Schematic of the \textit{Race car} model. Inputs are the agent's action $\mathbf{a}$, the gap time to the opponent $t_\mathrm{gap}$ and the additional lap time caused by the aerodynamic interaction $\Delta T_\mathrm{int}$. Output are the observation of the ego car $\mathbf{o}$ and the available observations for the opponent $\mathbf{\tilde o}$. For a detailed mathematical description, the reader is referred to \cite{fieni2025towards}.}
	\label{fig:racemodel}
\end{figure}

Given the current inputs and states of the system, the lap time is computed using lap time maps. For instance, allocating more battery energy or using a fresh soft compound results in faster lap times. On the other hand, deteriorated tires, a heavy car, or a pit stop increase the lap time. To account for the interaction between agents, we extend the lap time computation of \cite{fieni2025towards} with the interaction term as  
\begin{align}\label{eq:laptime}
T_\mathrm{lap} = & T_\mathrm{nom}(E_\mathrm{b}, \DeltaEb, \DeltaEf, m_\mathrm{car}, \PS)\nonumber\\
 &+\Delta T_{j}(\TW) + \Delta T_\mathrm{int}(t_\mathrm{gap}),
\end{align}
where $T_\mathrm{nom}$ is the nominal lap time map, $\Delta T_{j}$ is the additional time given by the tire wear and the chosen compound $j$, $\Delta T_\mathrm{int}$ is the additional lap time caused by the interaction, and $t_\mathrm{gap}$ the gap time. The latter two are introduced in \Cref{subsec:aeroint}.

Following the formulation of \cite{fieni2025towards}, the resulting states of the car are 
\begin{align}\label{eq:ssRL}
\mathbf{s} = & \left(\begin{matrix} E_\mathrm{b} & E_\mathrm{f} & m_\mathrm{car} & T_\mathrm{race} \end{matrix} \right.\ \dots \nonumber \\
                &  \left. \begin{matrix} b_\mathrm{cpd} & \TC & \TW & b_\mathrm{outlap} \end{matrix} \right),
\end{align}
where $b_\mathrm{cpd}$ and $b_\mathrm{outlap}$ are auxiliary variables indicating if at least two different compounds were employed (according to regulations) and if the previous lap was an outlap, respectively. These variables are needed to recover the Markov property \cite{fieni2025towards}. 

We distinguish between the observations of the ego car $\mathbf{o}$ and of the competitor $\mathbf{\tilde o}$. From the competitor's perspective, not all the states of the other car are completely observable. For instance, the battery level and its allocation strategy are not available information to other teams. Nevertheless, they can observe quantities such as the tire compound, the gap time, or if a pit stop is called. They can also track the tire age $\TA$ or if at least two different compounds were employed to meet the regulations. We choose the observation vector for the ego car as
\begin{equation}
    \mathbf{o} = \begin{pmatrix}
        \mathbf{s} & T_{\mathrm{lap}} & N_\mathrm{laps} - k
    \end{pmatrix},
\end{equation}
to inform the agent about the resulting lap time and the number of remaining laps. The observations available to the competitor are
\begin{equation}
    \mathbf{\tilde o} = \begin{pmatrix}
        \TA & \PS & b_\mathrm{cpd} & t_\mathrm{gap}
    \end{pmatrix}.
\end{equation}
Note that we choose the gap time $t_\mathrm{gap}$ instead of the lap time, although the latter is publicly available. This choice is motivated by two considerations. First, given the ego lap time and the gap time, the competitor's lap time is a redundant information. Second, the gap time is a derived variable that directly correlates with the additional lap time induced by the interaction. This facilitates the learning of the physical coupling. 

\subsection{Aerodynamic interaction model}\label{subsec:aeroint}
This model captures the physical coupling between agents. Slipstreaming reduces drag and downforce, accelerates tire degradation, and deteriorates brakes and engine cooling. These effects affect the lap time and influence the race strategy. In this work, only drag and downforce reduction are considered, although the framework can be extended to include additional interaction effects.

The dominance of drag or downforce depends on the characteristics of the circuit. On high-speed circuits such as the Autodromo Nazionale di Monza, drag reduction benefits the trailing car and can decrease lap times. Conversely, the Bahrain International Circuit features a lot of corners, and the downforce loss outweighs the drag reduction, resulting in slower laps. 

The game-theoretic framework presented in \cite{fieni2025game} enables accurate numerical investigations of the influence of slipstreaming conditions. By naturally capturing the underlying lap dynamics, this approach avoids the unnecessary overhead of explicit formulations. The lap time difference $\Delta T_\mathrm{int}$ for different initial $t_\mathrm{gap}$ is computed for the vehicle behind, subject to the wake effect of the leading car. We observed that for many circuits (not shown here) the underlying trend is well captured by a linear model:
\begin{align}
\Delta T_\mathrm{int} = 
\begin{cases}
a \cdot t_\mathrm{gap} + b, & \text{if } t_\mathrm{gap} \in [\qty{0.2}{\second}, \dots, \qty{1.35}{\second}], \\
0 & \text{else},
\end{cases}
\end{align}
where $a$ and $b$ are track-dependent fitting coefficients. \Cref{fig:fitInteraction} shows the resulting fitted trend for the Bahrain International Circuit considered in this paper. Where the function is non-zero, the values of $a$ and $b$ result in a lap time loss, i.e., being behind is detrimental in terms of lap time. 

\begin{figure}
	\centering
	\begin{externalize}{fitInteraction}
		\begin{tikzpicture}

\ifdefined\isMainDocument
  \def\datapath{figures/fitInteraction/}
  
  \def\width{\columnwidth}
  \def\height{0.4*\columnwidth}
  \def\voffset{1cm}
  
\else
  \def\datapath{}
  \def\width{15cm}
  \def\height{5cm}
  \def\voffset{2cm}
\fi

\begin{axis}[%
at={(0,0)},
width = \width,
height = \height,
xmin=0.1,
xmax=1.4,
xlabel=\small{$t_\mathrm{gap}$},
x unit =\unit{\second},
xtick={0.2,0.4,0.6,0.8,1,1.2},
ymin=0,
ymax=0.4,
ylabel=\small{$\Delta T_\mathrm{int}$},
y unit =\unit{\second},
ytick={0, 0.1,0.2,0.3,0.4},
axis background/.style={fill=white},
xmajorgrids,
ymajorgrids,
legend style={at={(0,0)}, anchor=south west, legend cell align=left, align=left, draw=white!15!black}
]
\addplot [color=black, line width=1.0pt]
  table[]{\datapath fitInteraction-1.tsv};
\addlegendentry{Numerical solution}

\addplot [color=black!50!white, dashed, line width=1.0pt]
  table[]{\datapath fitInteraction-2.tsv};
\addlegendentry{Fit}

\end{axis}
\end{tikzpicture}%
	\end{externalize}
	\caption{Additional lap time as a function of the gap time at the beginning of the lap. The numerical solution is derived from the framework in \cite{fieni2025game}.}
	\label{fig:fitInteraction}
\end{figure}

Finally, the gap times dynamics are modeled as 
\begin{align}
t_\mathrm{gap}^{1}[k+1] & = t_\mathrm{gap}^{1}[k] + (T_\mathrm{lap}^{1}[k] - T_\mathrm{lap}^{i}[k]),\\
t_\mathrm{gap}^{i}[k+1] & = t_\mathrm{gap}^{i}[k] + (T_\mathrm{lap}^{i}[k] - T_\mathrm{lap}^{1}[k]),
\end{align}
where $t_\mathrm{gap}>0$ indicates that the agent is behind. 

\subsection{Agent's architecture}\label{subsec:agent}
We introduce an agent structure to effectively handle the competitor's information, shown in \Cref{fig:agent}. The pre-trained \gls{sa} policy of \cite{fieni2025towards} is used as a backbone, whose neural network's layers are kept frozen during training. Its output is the nominal action $\mathbf{a}_\mathrm{nom}$. Inspired by \cite{mohseni2019interaction} and by Deepmind's $\alpha$-fold modular structure \cite{jumper2021highly}, we extend the agent with an interaction module, which outputs the action correction $\Delta \mathbf{a}$. The final policy is obtained by combining the outputs of the two modules as
\begin{equation}
\mathbf{a} = \mathbf{a}_\mathrm{nom} + \Delta \mathbf{a}.
\end{equation}
The \gls{sa} policy takes as input the observation vector $\mathbf{o}$, containing the complete state information of the ego vehicle.
To compute the correction term, the interaction module receives the same ego observation $\mathbf{o}$ and, in addition, the opponent’s information $\mathbf{\tilde o}$.

This architecture has two main advantages. First, the nominal policy already provides a robust baseline for \gls{sa} race strategies, as shown in \cite{fieni2025towards}. Second, the computational burden associated with training agents de novo is mitigated, stabilizing the training. The extended agent only requires fine-tuning to account for interaction effects, because only the \textit{correction} from the nominal policy has to be learned.

\begin{figure}
	\centering
	\begin{externalize}{Agent}
		\begin{tikzpicture}[scale=1, every node/.append style={outer sep=0pt}, >=stealth, font=\small]

\ifdefined\isMainDocument

\else
	\newcommand{\DeltaEb}{\Delta E_\mathrm{b,all}}
	\newcommand{\DeltaEf}{\Delta E_\mathrm{f,all}}
	\newcommand{\TW}{\mathrm{TW}}
	\newcommand{\TC}{\mathrm{TC}}
	\newcommand{\PS}{\mathrm{PS}}
	\newcommand{\bComp}{b_\mathrm{compound}}
\fi

\tikzstyle{block} = [draw, rectangle, minimum height=1cm, minimum width=1cm]
\def\dCirc{0.1cm}
\def\radius{0.3cm}
\tikzstyle{dot2} = [draw, circle, minimum size=\dCirc, fill = black, inner sep=0cm]
\tikzstyle{dot3} = [draw, circle, minimum size=\radius, color=white, fill = white, inner sep=0cm]

\def\altezza{2cm}
\def\larghezza{2.4cm}
\def\vertDist{2cm}
\def\horzDist{3.8cm}
\def\shadingOffset{0.06cm}
\def\textOffset{0.2cm}
\def\heightRef{0.07cm}
\def\widthRef{0.3cm}

\def\larghezzaTC{1.4*\larghezza}
\def\larghezzaTW{1.5*\larghezza}
\def\larghezzaLaptime{4.2*\larghezza}
\def\vertOffset{-\vertDist + \altezza}
\def\vertOffsetSecond{-2*\vertDist + \altezza}
\def\vertOffsetThird{-4*\vertDist + \altezza}
\def\oneThirdUp{\vertDist/3}
\def\twoThirdUp{2*\vertDist/3}

\node [coordinate] at (0.75*\larghezza,0) (Agent) {};
\node [block, fill=red!10, minimum height=2.5*\altezza, minimum width=2.7*\larghezza] at (Agent) (AgentBlock) {};
\node at ($(Agent) + (1.07*\larghezza, -1.12*\altezza)$) (TextAgent) {\textbf{Agent}};

\node [coordinate] at (0,0) (SAP) {};
\node [block, fill=gray!60, minimum height=\altezza, minimum width=\larghezza,align=center] at (SAP) (SAPBlock) {SA\\policy};
\node at ($(SAPBlock.south east)+(0,2.5pt)$) {
	\begin{tikzpicture}[scale=0.8,line width=1.1pt]
		\draw[rounded corners=0.5pt, fill=black] (0,0) rectangle (0.4,0.35);
		\draw (0.05,0.35) to (0.05, 0.5);
		\draw (0.05,0.5) arc[start angle=181, end angle=-1, radius=0.15];
		\draw (0.35,0.35) to (0.35, 0.5);
		\draw[fill=white] (0.2,0.2) circle (0.08);
	\end{tikzpicture}
};
\node [coordinate] at ($(SAPBlock) + (1.5*\larghezza, 0)$) (IM) {};
\node [block, fill=yellow!20, minimum height=\altezza, minimum width=\larghezza,align=center] at (IM) (IMBlock) {Interaction\\module};
\node at (IMBlock.south west) {
	\begin{tikzpicture}[scale=0.2, >=stealth]
		\draw[->, line width=0.5pt] (20:1) arc (20:190:1);
		\draw[->, line width=0.5pt] (200:1) arc (200:370:1);
	\end{tikzpicture}
};
\node [coordinate] at ($(AgentBlock.south) + (0,\vertDist/5)$) (Add) {};
\node [block, fill=white, minimum height=0, minimum width=0,align=center] at (Add) (AddBlock) {+};

\node [coordinate] at (SAPBlock.north) (sapi1) {};
\node [coordinate] at (SAPBlock.south) (sapo1) {};

\node [coordinate] at ($(IMBlock.north) + (-\larghezza/4,0)$) (imi1) {};
\node [coordinate] at ($(IMBlock.north) + (\larghezza/4,0)$) (imi2) {};
\node [coordinate] at (IMBlock.south) (imo1) {};

\node [coordinate] at (AddBlock.west) (addleft) {};
\node [coordinate] at (AddBlock.east) (addright) {};
\node [coordinate] at (AddBlock.south) (addbelow) {};

\node [coordinate] at ($(sapi1 |- AgentBlock.north)$) (o1) {};
\node [coordinate] at ($(imi2 |- AgentBlock.north)$) (o2) {};

\node [coordinate] at (AgentBlock.south) (a) {};

\node [coordinate] at ($(o1) + (0,-\vertDist/5)$) (belowo1) {};
\node [coordinate] at ($(belowo1 -| imi1)$) (aboveimi1) {};
\node [coordinate] at ($(sapo1 |- addleft)$) (belowsapo1) {};
\node [coordinate] at ($(imo1 |- addright)$) (belowimo1) {};

\draw[->] ($(o1)+(0,\vertDist/5)$) to node[midway, left, yshift=0.7cm, anchor=south east] {$\mathbf{o}$} (sapi1);
\draw[->] ($(o2)+(0,\vertDist/5)$) to node[midway, left, yshift=0.7cm, anchor=south east] {$\mathbf{\tilde o}$} (imi2);
\draw (belowo1) to (aboveimi1);
\draw[->] (aboveimi1) to (imi1);

\draw (sapo1) to node[midway, left] {$\mathbf{a}_\mathrm{nom}$} (belowsapo1);
\draw[->] (belowsapo1) to (addleft);

\draw (imo1) to node[midway, left] {$\Delta \mathbf{a}$} (belowimo1);
\draw[->] (belowimo1) to (addright);

\draw[->] (addbelow) to node[midway, left, yshift=-0.2cm] {$\mathbf{a}$} ($(a) + (0,-\vertDist/5)$);

\end{tikzpicture}
	\end{externalize}
	\caption{Schematic of the agent's structure. The \gls{sa} policy is taken from \cite{fieni2025towards} and its weights are kept frozen during training, while only the interaction module is trained. Inputs are the observation of the ego car $\mathbf{o}$ and the ones about the opponent $\mathbf{\tilde o}$. The nominal policy $\mathbf{a}_\mathrm{nom}$ is combined with the policy of the interaction module $\mathbf{\Delta a}$ to output the action $\mathbf{a}$.}
	\label{fig:agent}
\end{figure}

\subsection{Markov decision process}\label{subsec:markov}
The system is described by the state space  
\begin{equation}
\mathcal{S} = \begin{pmatrix} \mathbf{s}^{1} & t_\mathrm{gap}^{1} & \mathbf{s}^{i} & t_\mathrm{gap}^{i}\end{pmatrix} \in \mathbb{R}^{18},
\end{equation}
the observation space
\begin{equation}
\mathcal{O} = \begin{pmatrix} \mathbf{o}^{1} & \mathbf{\tilde o}^{i} \end{pmatrix} \in \mathbb{R}^{14}, 
\end{equation}
and the action space
\begin{equation}
\mathcal{A} = \begin{pmatrix} \mathbf{a}^{1} \end{pmatrix} \in \mathbb{R}^{3}, 
\end{equation}
such that $\mathcal{S}$, $\mathcal{O}$, $\mathcal{A}$ are feasible in the environment. The deterministic transition function 
\begin{equation}
T: \mathcal{S}\times\mathcal{A}\rightarrow\mathcal{S} \quad \text{such that} \quad \mathbf{s}_{k+1}' = T(\mathbf{s}_k', \mathbf{a}_k'), 
\end{equation}
with $\mathbf{s}'\in \mathcal{S}$ and $\mathbf{a}'\in \mathcal{A}$, captures the transition dynamics from lap $k$ to lap $k+1$. The episode terminates when no laps are left, i.e., $k = N_\mathrm{laps}$. 

The objective is to finish the race ahead of the opponent. However, a pure \textit{winner reward}, as in a zero-sum game, destabilizes training and suffers from reward sparsity. Additionally, it encourages detrimental behaviors, with agents focusing primarily on interfering with one another.
To promote reasonable and realistic strategies that account for interaction effects, we adopt the reward function
\begin{align}
R(\mathbf{s}_{k}',\mathbf{a}_{k}, \mathbf{s}_{k+1}') & = r_{k} \nonumber\\
& = r_\mathrm{step}+ r_\mathrm{final}.
\end{align}
It is composed by
\begin{equation}
r_\mathrm{step} = -t_\mathrm{gap}^{1}[k], \ \forall k \in [0,\dots, N_\mathrm{laps}],
\end{equation}
where $t_\mathrm{gap}$ is numerically well-scaled and the negative sign is used to promote being ahead, and
\begin{equation}
r_\mathrm{final} = 
\begin{cases}
c_\mathrm{win}, &\text{if } t_\mathrm{gap}^{1}[N_\mathrm{laps}]< 0, \\
0, &\text{else,}
\end{cases}
\end{equation} 
where $c_\mathrm{win}$ is a constant winner reward, chosen to be approximately in the same order of magnitude as the cumulated step reward. Maximizing the total reward therefore corresponds to the similar goals of winning and increasing the distance from the opponent, without the drawbacks of sparse reward functions. This results in a non-zero-sum game. The discount factor is set to $\gamma = 1$. 

Each module within the agent architecture is trained with a distinct reward function aligned with its specific scope. While the underlying \gls{sa} policy is trained to minimize race time, the interaction module is dedicated to handle the interaction with the opponent. This modular training ensures that the agent achieves a competitive baseline performance while managing multi-agent dynamics.

Since the process satisfies the Markov property, by definition the transition probability satisfies
\begin{equation}
\mathbb{P}\big(\mathbf{s}_{k+1} \mid \mathbf{s}_0, \dots, \mathbf{s}_k,
\mathbf{a}_0, \dots, \mathbf{a}_k\big)
=
\mathbb{P}\big(\mathbf{s}_{k+1} \mid \mathbf{s}_k, \mathbf{a}_k\big),
\end{equation}
for all $k$. Finally, the finite-horizon \gls{mdp} 
\begin{equation}
\mathcal{M} = (\mathcal{S}, \, \mathcal{O},\, \mathcal{A},\, T,\, R, \,\mathbb{P}, \,\gamma)
\end{equation}
formalizes the race strategy optimization problem considering the competitor's interaction.

\section{Multi-agent extension}\label{sec:MAextension}
In this section, we describe the generation of multiple agents for the \textit{battle arena}. A ranking system is introduced to assess and order agents' performance, and the agents are trained employing a custom self-play training scheme.

\subsection{Ranking system}\label{subsec:rankingsystem}
Agents with similar rewards may have different performance, because the reward function balances cumulative race time with winner reward. Thus, an agent with suboptimal pace may still win under favorable initial conditions, while a faster agent facing adverse conditions may lose. 

We adopt the Elo rating system \cite{elo1978rating} to measure the policies' relative performance. After each match, ratings are updated based on the difference between the players’ rankings: Defeating a stronger opponent yields a larger increase than defeating a weaker one. This mechanism is more representative of the relative performance than the winning probability, since it prevents agents from achieving high rankings only by outperforming weak opponents.

We use the Elo rating during and after training, since it can be continuously updated. In the former case, to monitor the performance of the agents and to select the pool of opponents. In the latter case, the $m$ highest-ranked agents compete until their ratings converge, creating a \textit{battle arena} that orders agents by performance. This methodology allows new agents to be integrated later. 

\subsection{Self-play training}\label{subsec:training}
Training methods for multi-agent systems are computationally demanding. Common approaches include centralized and decentralized multi-agent \gls{rl}. The former is well suited for cooperative tasks, as agents share their experiences, whereas the latter is more appropriate for competitive settings, where agents learn independently. 

Self-play \cite{silver2017mastering,silver2018general} provides an efficient and effective framework for training structurally identical agents. One agent interacts with the environment while competing against an identical, fixed opponent embedded in it. During training, only the learning agent is updated. After convergence, the opponent is replaced with the newly trained agent, and a new self-play iteration begins. The policy is progressively improved by competing always against an opponent with similar expertise. 

In the environment shown in \Cref{fig:environment}, \textit{Agent 1} is the learning agent, whereas \textit{Agent i} is the opponent updated at each self-play iteration. Although this procedure breaks the Markov property across iterations, it still holds within each training episode. 

Combining self-play with the interaction module improves learning efficiency. First, only the interaction module is updated, refining the already trained \gls{sa} policy, which maintains the baseline strategy. Second, training is performed on a single agent at a time, avoiding the overhead associated with decentralized training of multiple agents.  

\subsubsection*{Custom self-play} To diversify training and encourage exploration, we modify the standard self-play architecture as illustrated in \Cref{fig:selfplay}. The procedure is structured as follows.
\begin{enumerate}
\item During the initial self-play iteration, the opponent is restricted to the pre-trained \gls{sa} policy, preventing the learning of strategies based on an uninitialized interaction module. 
\item When the training agent's Elo rating exceeds that of the current best agent by 10 points, its policy is saved and added to the opponent pool. To prevent the generation of agents while this condition is continuously met, a cooldown period of 50 episodes is enforced before a new policy can be saved. 
\item To promote competitive policies, only the currently top-ranked $m$ agents are eligible to be selected as opponents. 
\item To mitigate overfitting to a specific adversarial strategy, opponents are not held static until convergence. Instead, at each episode an opponent is uniformly sampled at random from the active pool. 
\item The training is terminated when the learning agent demonstrates no further improvement against any of the current top-ranked $m$ agents. 
\end{enumerate}
\begin{figure}
	\centering
	\scalebox{0.9}{
	\begin{externalize}{SelfPlay}
		\begin{tikzpicture}[scale=1, every node/.append style={outer sep=0pt}, >=stealth, font=\small]

\ifdefined\isMainDocument

\else
	\newcommand{\DeltaEb}{\Delta E_\mathrm{b,all}}
	\newcommand{\DeltaEf}{\Delta E_\mathrm{f,all}}
	\newcommand{\TW}{\mathrm{TW}}
	\newcommand{\TC}{\mathrm{TC}}
	\newcommand{\PS}{\mathrm{PS}}
	\newcommand{\bComp}{b_\mathrm{compound}}
\fi

\tikzstyle{block} = [draw, rectangle, minimum height=1cm, minimum width=1cm]
\def\dCirc{0.1cm}
\def\radius{0.3cm}
\tikzstyle{dot2} = [draw, circle, minimum size=\dCirc, fill = black, inner sep=0cm]
\tikzstyle{dot3} = [draw, circle, minimum size=\radius, color=white, fill = white, inner sep=0cm]

\def\altezza{1cm}
\def\larghezza{2cm}
\def\vertDist{2cm}
\def\horzDist{3.8cm}
\def\shadingOffset{0.06cm}
\def\textOffset{0.2cm}
\def\heightRef{0.07cm}
\def\widthRef{0.3cm}

\def\larghezzaAgent{1.1*\larghezza}
\def\larghezzaTW{1.4*\larghezza}
\def\larghezzaLaptime{4.2*\larghezza}
\def\vertOffset{-\vertDist + \altezza}
\def\vertOffsetSecond{-2*\vertDist + \altezza}
\def\vertOffsetThird{-4*\vertDist + \altezza}
\def\oneThirdUp{\vertDist/3}
\def\twoThirdUp{2*\vertDist/3}

\node [coordinate] at (0,0) (agent11) {};
\node [block, fill=red!10, minimum height=\altezza, minimum width=\larghezzaAgent] at (agent11) (Agent11Block) {Agent$_{1,1}$};
\node at (Agent11Block.south east) {
	\begin{tikzpicture}[scale=0.15, >=stealth]
		\draw[->, line width=0.5pt] (20:1) arc (20:190:1);
		\draw[->, line width=0.5pt] (200:1) arc (200:370:1);
	\end{tikzpicture}
};

\node [coordinate] at ($(agent11) + (0,-\vertDist)$) (agent1n) {};
\node [block, fill=red!10, minimum height=\altezza, minimum width=\larghezzaAgent] at (agent1n) (Agent1nBlock) {Agent$_{1,\mathrm{best}}$};
\node at (Agent1nBlock.south east) {
	\begin{tikzpicture}[scale=0.15, >=stealth]
		\draw[->, line width=0.5pt] (20:1) arc (20:190:1);
		\draw[->, line width=0.5pt] (200:1) arc (200:370:1);
	\end{tikzpicture}
};

\node [coordinate] at (0,-0.5*\vertDist) (iter1) {};
\node [block, dotted, minimum height=3.5*\altezza, minimum width=1.3*\larghezzaAgent] at (iter1) (Iter1Block) {};
\node at ($(Iter1Block.north west) + (0.45cm,0.15cm)$) {Iter. 1};

\node [coordinate] at ($(agent1n) + (0,-1.25*\vertDist)$) (agent21) {};
\node [block, fill=red!10, minimum height=\altezza, minimum width=\larghezzaAgent] at (agent21) (Agent21Block) {Agent$_{n,1}$};
\node at (Agent21Block.south east) {
	\begin{tikzpicture}[scale=0.15, >=stealth]
		\draw[->, line width=0.5pt] (20:1) arc (20:190:1);
		\draw[->, line width=0.5pt] (200:1) arc (200:370:1);
	\end{tikzpicture}
};

\node [coordinate] at ($(agent21) + (0,-\vertDist)$) (agent2m) {};
\node [block, fill=red!10, minimum height=\altezza, minimum width=\larghezzaAgent] at (agent2m) (Agent2mBlock) {Agent$_{n,\mathrm{best}}$};
\node at (Agent2mBlock.south east) {
	\begin{tikzpicture}[scale=0.15, >=stealth]
		\draw[->, line width=0.5pt] (20:1) arc (20:190:1);
		\draw[->, line width=0.5pt] (200:1) arc (200:370:1);
	\end{tikzpicture}
};

\node [coordinate] at (0,-2.75*\vertDist) (iter2) {};
\node [block, dotted, minimum height=3.5*\altezza, minimum width=1.3*\larghezzaAgent] at (iter2) (Iter2Block) {};
\node at ($(Iter2Block.north west) + (0.45cm,0.15cm)$) {Iter. $n$};

\node [coordinate] at (1.5*\larghezzaAgent,-0.5*\vertDist) (env1) {};
\node [block, fill=gray!20, minimum height=3.5*\altezza, minimum width=1.3*\larghezzaAgent] at (env1) (Env1Block) {};
\node at ($(Env1Block.north west) + (0.9cm,0.15cm)$) {Environment};

\node [coordinate] at (1.5*\larghezzaAgent,-2.75*\vertDist) (env2) {};
\node [block, fill=gray!20, minimum height=3.5*\altezza, minimum width=1.3*\larghezzaAgent] at (env2) (Env2Block) {};

\node [coordinate] at (1.5*\larghezzaAgent,-0.5*\vertDist) (agent11lock) {};
\node [block, fill=red!10, minimum height=\altezza, minimum width=\larghezzaAgent,align=center] at (agent11lock) (Agent11LockBlock) {SA\\policy};
\node at ($(Agent11LockBlock.south west)+(0,2.5pt)$) {
	\begin{tikzpicture}[scale=0.8,line width=1.1pt]
		\draw[rounded corners=0.5pt, fill=black] (0,0) rectangle (0.4,0.35);
		\draw (0.05,0.35) to (0.05, 0.5);
		\draw (0.05,0.5) arc[start angle=181, end angle=-1, radius=0.15];
		\draw (0.35,0.35) to (0.35, 0.5);
		\draw[fill=white] (0.2,0.2) circle (0.08);
	\end{tikzpicture}
};

\node [coordinate] at (1.5*\larghezzaAgent,-2.75*\vertDist) (opponent) {};
\node [block, fill=red!10, minimum height=\altezza, minimum width=\larghezzaAgent] at (opponent) (Opponent) {Opponent};
\node at ($(Opponent.south west)+(0,2.5pt)$) {
	\begin{tikzpicture}[scale=0.8,line width=1.1pt]
		\draw[rounded corners=0.5pt, fill=black] (0,0) rectangle (0.4,0.35);
		\draw (0.05,0.35) to (0.05, 0.5);
		\draw (0.05,0.5) arc[start angle=181, end angle=-1, radius=0.15];
		\draw (0.35,0.35) to (0.35, 0.5);
		\draw[fill=white] (0.2,0.2) circle (0.08);
	\end{tikzpicture}
};

\node [coordinate] at (2.85*\larghezzaAgent,-2.25*\vertDist) (agent1alock) {};
\node [block, fill=red!10, minimum height=\altezza, minimum width=\larghezzaAgent] at (agent1alock) (Agent1aLockBlock) {Agent$_{1,\mathrm{best}}$};
\node[anchor=center] at (2.85*\larghezzaAgent,-2.71*\vertDist) (agent1block) {\textbf{\vdots}};

\node [coordinate] at (2.85*\larghezzaAgent,-3.25*\vertDist) (agent1clock) {};
\node [block, fill=red!10, minimum height=\altezza, minimum width=\larghezzaAgent] at (agent1clock) (Agent1cLockBlock) {Agent$_{n-1,\mathrm{best}}$};
\node [coordinate] at ($(Iter1Block.east) + (0,\vertDist/2)$) (iter1o) {};
\node [coordinate] at ($(Iter1Block.east) + (0,-\vertDist/2)$) (iter1i) {};

\node [coordinate] at ($(Iter2Block.east) + (0,\vertDist/2)$) (iter2o) {};
\node [coordinate] at ($(Iter2Block.east) + (0,-\vertDist/2)$) (iter2i) {};

\node [coordinate] at ($(Env1Block.west) + (0,\vertDist/2)$) (env1i) {};
\node [coordinate] at ($(Env1Block.west) + (0,-\vertDist/2)$) (env1o) {};

\node [coordinate] at ($(Env2Block.west) + (0,\vertDist/2)$) (env2i) {};
\node [coordinate] at ($(Env2Block.west) + (0,-\vertDist/2)$) (env2o) {};

\node [coordinate] at ($(Agent1nBlock.south) + (0,-\vertDist/4)$) (belowiter) {};
\node [coordinate] at ($(belowiter -| Opponent.north)$) (aboveagent1n) {};

\draw[->,dashed] (Agent11Block.south) to node[midway, left, align=center,font=\tiny] {Training} (Agent1nBlock.north);
\draw[->] (iter1o) to node[midway, above, align=center] {$\mathbf{a}$} (env1i);
\draw[->] (env1o) to node[midway, above, align=center] {$\mathbf{o}$} (iter1i);

\draw (Agent1nBlock.south) to ($(Agent1nBlock.south) + (0,-0.25*\vertDist)$);
\node[anchor=center] at ($(Agent1nBlock.south) + (0,-0.33*\vertDist)$) (dots) {\textbf{\vdots}};
\draw[->] ($(Agent1nBlock.south) + (0,-0.5*\vertDist)$) to (Agent21Block.north);

\draw[->,dashed] (Agent21Block.south) to node[midway, left, align=center,font=\tiny] {Training} (Agent2mBlock.north);
\draw[->] (iter2o) to node[midway, above, align=center] {$\mathbf{a}$} (env2i);
\draw[->] (env2o) to node[midway, above, align=center] {$\mathbf{o}$} (iter2i);

\draw[->] (Opponent.east) .. controls ($(Env1Block.north east |- Agent1aLockBlock)$) .. (Agent1aLockBlock.west);
\draw[->] (Opponent.east) .. controls ($(Env2Block.south east |- Agent1cLockBlock)$) .. (Agent1cLockBlock.west);

\end{tikzpicture}
	\end{externalize}
	}
	\caption{Custom self-play training scheme. The training agent is shown on the left, while the opponent embedded in the environment remains fixed. During the first iteration, the \gls{sa} policy is the only opponent. After each iteration $n$, the training agent with the highest Elo score (``best''), is added to the pool of potential future opponents of iteration $n+1$.}
	\label{fig:selfplay}
\end{figure}

\subsubsection*{Training details} Before each episode, the initial gap time is randomly sampled to promote policy robustness against varying initial conditions. Although the agent was trained for $12000$ episodes, the proposed self-play mechanism demonstrated convergence within the first $2000$ episodes. The policy was optimized using a standard \gls{sac} algorithm, and the training on a commercial workstation (Apple M2 Max, \qty{32}{\giga\byte} RAM) requires approximately \qty{3}{\hour}.

\section{Results}\label{sec:results}
In this section, we showcase the validity of the presented framework. First, we validate the superiority of the interaction module against the \gls{sa} strategy. Then, we analyze a race between two agents with an \textit{undercut}, a typical tactical maneuver employed in \gls{f1} races. In a last case study, we show how the agents adapt their strategy to responding opponents.

\Cref{tab:ranking} summarizes the converged Elo score and ranking position for the $m=4$ best agents generated during training. They constitute the considered \textit{battle arena}.
\begin{table}
\begin{center}
\caption{Elo score and ranking position for the agents of the considered battle arena with $m=4$.}\label{tab:ranking}
\begin{tabular}{c c c c c}
\toprule
\textbf{Agent} & $A$ & $B$ & $C$ & $D$\\
\midrule
\textbf{Elo score} & 1970 & 946 & 914 & 970\\
\midrule
\textbf{Ranking} & $1^{\circ}$ & $3^{\circ}$ & $4^{\circ}$ & $2^{\circ}$\\
\bottomrule
\end{tabular}
\end{center}
\end{table}
To define the pit stop strategy we use the notation $(\TC_{k},\dots)$, where $\TC\in\{S,M,H\}$ is the tire compound ($S$ stands for ``soft'', $M$ for ``medium'' and $H$ for ``hard''), and $k$ is the lap where that compound is mounted.

The system has to be initialized. To allow for fair comparison, every agent starts on medium tires. Except for \Cref{subsec:RLCS1}, agent $A$ always starts the race $\qty{0.5}{\second}$ behind the other agents, simulating a realistic race start. 

\subsection{Ablation study - Superiority of the interaction module}\label{subsec:RLCS1}
To isolate the positive effect of the interaction module, we perform an ablation study by considering the two setups shown in \Cref{fig:setupablation}. In both of them, the agent behind experiences the additional lap time caused by the interaction. In the first setup, the \gls{sa} policy competes against itself. The race time achieved by the agent starting behind is denoted as $T_\mathrm{race,SA}$. The second setup consists of the \gls{sa} policy starting ahead, and an agent \textit{with} interaction module starting behind. The race time of this latter is denoted as $T_\mathrm{race,MA}$. To assess the performance of the interaction module, we compute the race time advantage 
\begin{equation}
\Delta T_\mathrm{race,adv} = T_\mathrm{race,MA} - T_\mathrm{race,SA}.
\end{equation}
A negative $\Delta T_\mathrm{race,adv}$ means that the agent with the interaction module improved its race time compared to the \gls{sa} policy for the same initial gap. 

 \begin{figure}
	\centering
	\begin{externalize}{SetupCS11}
		\begin{tikzpicture}[scale=1, every node/.append style={outer sep=0pt}, >=stealth, font=\small]

\ifdefined\isMainDocument

\else
	\newcommand{\DeltaEb}{\Delta E_\mathrm{b,all}}
	\newcommand{\DeltaEf}{\Delta E_\mathrm{f,all}}
	\newcommand{\TW}{\mathrm{TW}}
	\newcommand{\TC}{\mathrm{TC}}
	\newcommand{\PS}{\mathrm{PS}}
	\newcommand{\bComp}{b_\mathrm{compound}}
\fi

\tikzstyle{block} = [draw, rectangle, minimum height=1cm, minimum width=1cm]
\def\dCirc{0.1cm}
\def\radius{0.3cm}
\tikzstyle{dot2} = [draw, circle, minimum size=\dCirc, fill = black, inner sep=0cm]
\tikzstyle{dot3} = [draw, circle, minimum size=\radius, color=white, fill = white, inner sep=0cm]

\def\altezza{1cm}
\def\larghezza{2cm}
\def\vertDist{2.5cm}
\def\horzDist{3.8cm}
\def\shadingOffset{0.06cm}
\def\textOffset{0.2cm}
\def\heightRef{0.07cm}
\def\widthRef{0.3cm}

\def\larghezzaAgent{1.1*\larghezza}
\def\larghezzaTW{1.4*\larghezza}
\def\larghezzaLaptime{4.2*\larghezza}
\def\vertOffset{-\vertDist + \altezza}
\def\vertOffsetSecond{-2*\vertDist + \altezza}
\def\vertOffsetThird{-4*\vertDist + \altezza}
\def\oneThirdUp{\vertDist/3}
\def\twoThirdUp{2*\vertDist/3}

\node [coordinate] at (0,0) (agent1) {};
\node [block, fill=red!10, minimum height=\altezza, minimum width=\larghezzaAgent,align=center] at (agent1) (SApolicy) {SA\\policy};

\node [coordinate] at (\horzDist,0) (agent2) {};
\node [block, fill=red!10, minimum height=\altezza, minimum width=\larghezzaAgent,align=center] at (agent2) (SApolicy1) {SA\\policy};

\draw[<-] (SApolicy.east) to node[midway, above, align=center] {$t_\mathrm{gap,init}$} (SApolicy1.west);
\node[anchor=north] at (SApolicy1.south) {$T_\mathrm{race,SA}$};

\node [coordinate] at (0,-\vertDist) (agent3) {};
\node [block, fill=red!10, minimum height=\altezza, minimum width=\larghezzaAgent,align=center] at (agent3) (SApolicy2) {SA\\policy};

\node [coordinate] at (\horzDist,-\vertDist) (agent4) {};
\node [block, fill=red!10, minimum height=\altezza, minimum width=\larghezzaAgent,align=center] at (agent4) (MApolicy) {Agents \\ $A$, $B$, $C$, $D$};

\draw[<-] (SApolicy2.east) to node[midway, above, align=center] {$t_\mathrm{gap,init}$} (MApolicy.west);
\node[anchor=south] at (MApolicy.north) {$T_\mathrm{race,MA}$};

\end{tikzpicture}
	\end{externalize}
	\caption{Setups for the ablation study. Above, the \gls{sa} policy competes against itself. The race time achieved by the agent starting behind is recorded as $T_\mathrm{race,SA}$. Below, the agents $A$, $B$, $C$, and $D$ start behind and compete against the \gls{sa} policy. Their achieved race time is recorded as $T_\mathrm{race,MA}$.}
	\label{fig:setupablation}
\end{figure}
\Cref{fig:CS11} presents $\Delta T_\mathrm{race,adv}$ for the considered agents and different initial gap times. Each agent consistently achieves a better race time than the \gls{sa} policy when racing behind, with the improvement aligned with its ranking of \Cref{tab:ranking}. This demonstrates the added value provided by the interaction module trained via self-play, which explicitly accounts for the aerodynamic interaction and responds with a better strategy. 

Despite not being trained specifically for the aerodynamic interaction, the \gls{sa} policy reacts to disturbances by adapting the strategy (as shown in \cite{fieni2025towards}). However, it does not \textit{actively} respond to the actions of the other agents. Since the environment corresponds to a game, agents with interaction module are expecting an \textit{active} response from the \gls{sa} policy. These considerations further strengthens the race time improvement achieved by agents with interaction module.
 \begin{figure}
	\centering
	\begin{externalize}{CS11}
		\begin{tikzpicture}

\ifdefined\isMainDocument
  \def\datapath{figures/CS11/}
  
  \def\width{\columnwidth}
  \def\height{0.8*\columnwidth}
  
\else
  \def\datapath{}
  \def\width{15cm}
  \def\height{8cm}
\fi

\begin{axis}[%
width=\width,
height=\height,
at={(0,0)},
xmin=0,
xmax=2.6,
xlabel={Initial gap time},
x unit=\unit{s},
xtick = {0, 0.2, 0.4, 0.6, 0.8, 1, 1.2, 1.4, 1.6, 1.8, 2, 2.2, 2.4},
xticklabels = {0, , 0.4, , 0.8, , 1.2, , 1.6, , 2, , 2.4},
ymin=-20,
ymax=-11,
ylabel={$\Delta T_\mathrm{race,adv}$},
y unit=\unit{s},
ytick = {-12, -14, -16, -18, -20},
axis background/.style={fill=white},
xmajorgrids,
ymajorgrids,
legend style={at={(1,1)}, legend cell align=left, align=left, draw=white!15!black}
]
\addplot [color=black, dotted, line width=1.0pt, mark size=2.0pt, mark=*, mark options={solid, fill=white, black}]
  table[]{\datapath T_raceAdv-1.tsv};
\addlegendentry{A}

\addplot [color=black, dotted, line width=1.0pt, mark size=3.0pt, mark=x, mark options={solid, black}]
  table[]{\datapath T_raceAdv-3.tsv};
\addlegendentry{B}

\addplot [color=black, dotted, line width=1.0pt, mark size=2pt, mark=diamond*, mark options={solid, fill=white, black}]
  table[]{\datapath T_raceAdv-4.tsv};
\addlegendentry{C}

\addplot [color=black, dotted, line width=1.0pt, mark size=3.0pt, mark=+, mark options={solid, black}]
  table[]{\datapath T_raceAdv-2.tsv};
\addlegendentry{D}

\end{axis}

\end{tikzpicture}%
	\end{externalize}
	\caption{$\Delta T_\mathrm{race,adv}$ for the agents considered and different initial gap times.}
	\label{fig:CS11}
\end{figure}

The advantage of the interaction module is not only given by the superiority in the race time, but rather to exploit tactical advantages, as we will show in \Cref{subsec:RLCS2}. 

\subsection{Battle between two agents with undercut}\label{subsec:RLCS2}
\Cref{fig:CS1} shows the fuel energy allocation, the battery energy allocation, the pit stops decisions, and the race time difference for $A$ against $B$. In particular, $A$ chooses a $(M_{0},S_{19},S_{33})$ strategy while $B$ goes for $(M_{0},S_{22},S_{49})$.
\begin{figure}
	\centering
	\begin{externalize}{CS1}
		\begin{tikzpicture}

\ifdefined\isMainDocument
  \def\datapath{figures/CS1/}
  
  \def\width{\columnwidth}
  \def\height{0.4*\columnwidth}
  \def\voffset{1cm}
  
\else
  \def\datapath{}
  \def\width{15cm}
  \def\height{5cm}
  \def\voffset{2cm}
\fi

\begin{axis}[%
at={(0,0)},
width = \width,
height = \height,
xmin=0,
xmax=56,
xtick={0,10,19,20,22,30,33,39,40,49,50},
xticklabels={{},{},{},{},{},{},{},{},{},{},{}},
ylabel=\small{$\Delta E_\mathrm{f,all}$},
y unit =-,
ytick = {90, 100, 110},
yticklabels = {90\%, 100\%, 110\%},
ymin=85,
ymax=115,
axis background/.style={fill=white},
xmajorgrids,
ymajorgrids,
legend style={at={(1,1)}, anchor=north east, legend cell align=left, align=left, draw=white!15!black}
]
\addplot [color=blue, line width=1.0pt]
  table[]{\datapath CS1-1.tsv};
\addlegendentry{Agent A}

\addplot [color=red, line width=1.0pt]
  table[]{\datapath CS1-2.tsv};
\addlegendentry{Agent B}

\addplot [color=black, dashed, line width=0.5pt, forget plot]
  table[]{\datapath CS1-3.tsv};
\addplot [color=black, dashed, line width=0.5pt, forget plot]
  table[]{\datapath CS1-4.tsv};
\end{axis}

\begin{axis}[%
at={(0,-0.6*\height)},
width = \width,
height = \height,
xmin=0,
xmax=56,
xtick={0,10,19,20,22,30,33,39,40,49,50},
xticklabels={{},{},{},{},{},{},{},{},{},{},{}},
ylabel=\small{$\Delta E_\mathrm{b,all}$},
y unit=-,
ymin=-1.3,
ymax=1.3,
axis background/.style={fill=white},
xmajorgrids,
ymajorgrids
]
\addplot [color=blue, line width=1.0pt, forget plot]
  table[]{\datapath CS1-5.tsv};
\addplot [color=red, line width=1.0pt, forget plot]
  table[]{\datapath CS1-6.tsv};
\addplot [color=black, dashed, line width=0.5pt, forget plot]
  table[]{\datapath CS1-7.tsv};
\addplot [color=black, dashed, line width=0.5pt, forget plot]
  table[]{\datapath CS1-8.tsv};
\end{axis}

\begin{axis}[%
at={(0,-1.2*\height)},
width = \width,
height = \height,
xmin=0,
xmax=56,
xtick={0,10,19,20,22,30,33,39,40,49,50},
xticklabels={{},{},{},{},{},{},{},{},{},{},{}},
xlabel={},
ymin=-0.3,
ymax=3.3,
ytick={0,1,2,3},
yticklabels={No pit, Pit for $S$, Pit for $M$, Pit for $H$},
axis background/.style={fill=white},
xmajorgrids,
ymajorgrids
]
\addplot [ycomb, color=blue, mark=o, mark size=1.5pt, line width=0.5pt]
  table[]{\datapath CS1-9.tsv};
\addplot [ycomb, color=red, mark=o, mark size=2.5pt, line width=0.5pt]
  table[]{\datapath CS1-10.tsv};
\end{axis}

\begin{axis}[%
at={(0,-1.8*\height)},
width = \width,
height = \height,
xmin=0,
xmax=57,
xtick={0,10,19,20,22,30,33,39,40,49,50},
xticklabels={0,10,{},20,{},30,{},{},40,{},50},
xlabel={Lap},
ytick={-30, -20, -10, 0, 10, 20, 30},
ymin=-33,
ymax=33,
ylabel={$\Delta T$},
y unit=\unit{\second},
axis background/.style={fill=white},
xmajorgrids,
ymajorgrids,
legend style={at={(0,0)}, anchor=south west, legend cell align=left, align=left, draw=white!15!black}
]
\addplot [color=black, line width=1.0pt]
  table[]{\datapath CS1-11.tsv};
\addlegendentry{$\Delta T<0 \rightarrow$ A is ahead.}

\end{axis}

\end{tikzpicture}%
	\end{externalize}
	\caption{Race strategies and race time difference for the duel between $A$ (in blue) and $B$ (in red). The first plot shows the normalized fuel energy allocation, the second one the normalized battery energy allocation, and the third one the pit stop decision variable. Agent $A$ starts $\qty{0.5}{\second}$ behind $B$, and a negative gap time means that agent $A$ is ahead.}
	\label{fig:CS1}
\end{figure}

For the first third of the race, $A$ remains behind $B$. Until the first pit stop, $A$ maintains a gap of approximately $\qty{1.6}{s}$, staying outside the wake effect region that would otherwise increase the lap time. The fuel energy allocation strategy substantially differs, with $A$ allocating less fuel per lap than $B$.   

Thanks to the first pit stop phase, $A$ performs a successful \textit{undercut}. After pitting on lap 19, $A$ rejoins the track outside the wake of $B$. With fresh soft tires and exploiting the previously conserved fuel, $A$ reduces the gap. Indeed, when $B$ pits at lap 22, it rejoins behind. This tactical maneuver is employed when racing behind a competitor is detrimental in terms of lap time. Thanks to the interaction module, agent $A$ shows that this strategy directly emerges from the fact that our setup explicitly accounts for multi-agent interactions.

Just before being overtaken, $A$ pits again at lap 33, avoiding the disadvantage of the aerodynamic interaction. Thanks to fresher tires and the lighter car, $A$ closes the gap to $B$ over the subsequent 17 laps, while $B$ increasingly suffers from tire degradation effects. At this point, $A$ secured the race lead, as it will remain ahead regardless of $B$'s strategy. Nevertheless, $B$ decides to pit anyway due to severe tire wear (soft tires with an age of 27 laps), prioritizing a competitive race time.

$A$ wins the race by $\qty{12.82}{\second}$. Balancing tire degradation with fuel allocation, $A$ prevails over $B$ over a race lasting approximately $\qty{90}{\minute}$. We observe how both agents react to the current race state, responding to the competitor's actions and observing its behavior. 

\subsection{Adapting the strategy}\label{subsec:RLCS3}
To further illustrate the agent response to different opponents, we evaluate pairwise races in which two agents compete simultaneously. The resulting pit stop strategies are shown in \Cref{fig:CS2}. Race times are compared against a \textit{constant} baseline defined as the race time of $A$ when competing against $C$
\begin{equation}
\Delta T_\mathrm{race} = T_\mathrm{race,i}(N_\mathrm{laps}) - T_\mathrm{Baseline},
\end{equation}
and are displayed on the right $y$-axis of \Cref{fig:CS2}.
\begin{figure}
	\centering
	\begin{externalize}{CS2}
		\input{figures/CS2/CS2.tex}
	\end{externalize}
	\caption{Pit stop and tire compound strategies for duels between different agents. Yellow indicates the medium compound and red the soft compound. Tire icons mark the pit stop laps. In each pair, the agent shown above starts $\qty{0.5}{\second}$ behind the agent below. On the right side, the race time differences are shown, all computed with respect to a common baseline, defined by agent $A$ competing against $C$.}
	\label{fig:CS2}
\end{figure}

Agent $A$ wins against every opponent. While $B$, $C$, and $D$ have comparable Elo scores, the difference of 1000 points indicates a clear dominance of $A$, even with different initial gap times (not shown). It is important to emphasize that energy management constitutes a critical component of race strategy, and race outcomes are not only determined by pit-stop decisions. 

The policy of $A$ is distinguished by its robustness. Across all duels, $A$ consistently adopts a two-stop strategy $(M,S,S)$, modifying the pit-stop laps to react to the opponent. This behavior indicates that the interaction module works as intended. Indeed, variations in the pit stop timing are performed to react to the competitor's response, while preserving the underlying strategy. The energy management is adjusted accordingly, although omitted here for brevity. Although $B$, $C$, and $D$ explore alternative strategies, they do not result in successful ones. 

The agents' race times correlate with their ranking, with higher-ranked agents consistently achieving lower total race times. Even among different games, all agents remain consistent in their own race times, even when using different strategies. Regardless of the opponent, the $\Delta T_\mathrm{race}$ of $A$ remains on the same order of magnitude. Similar consistency is observed for $B$ and $C$, whose race time vary only marginally when racing against different opponents.

Except for $A$, the other agents change their strategies depending on the opponent. For instance, when $C$ competes against $A$ and starts ahead, it adopts a three-stops strategy, whereas when starting behind $B$ it switches to a two-stops strategy. Similarly, $B$ performs two stops when starting ahead of $A$, but reduces to a single stop when racing against $C$. From a race strategist perspective, selecting among multiple strategies with comparable race times remains challenging. This further motivates the use of \gls{rl} agents to identify robust choices.

In general, all the agents show a clear preference for the soft compound. While this choice aligns with common practice at the Bahrain International Circuit, it strongly depends on the tire degradation model and on the pre-trained \gls{sa} policy. Additionally, enforcing the initial compound might condition the preference towards certain strategies. Diversification of training on initial compounds and further investigations could reveal new strategies.

\section{Conclusions}\label{sec:conclusions}
In this paper, we presented a framework for training \gls{rl} agents that determines \gls{f1} race strategies while accounting for interactions with competitors. Each agent decides the energy management, pit-stop timing, and tire compound selection. Building upon a \gls{sa} policy, we extended the agent by introducing an interaction module to react to opponents’ behavior. The combination of this module with a self-play training scheme is crucial for efficient learning of multi-agent settings. Moreover, the agent's backbone pre-trained policy retains the robust race time performance. 

The proposed approach generates agents with distinct performance levels and strategic behaviors. Results show that they adapt their strategy based on the current race state and competitors' actions. As the agents rely exclusively on information available during real races, this tool can support the decision-making of race strategists, complementing the large number of Monte Carlo simulations typically required. 

Future work could incorporate the inclusion of traffic through probabilistic representations, as well as additional stochastic factors such as safety cars or weather predictions. The learned policies could also be analyzed to identify potential Nash equilibria in strategic interactions. Concerning training, opponents' labeling may improve the performance, by allowing the interaction module to tailor responses to specific competitors.

\bibliographystyle{IEEEtran}
\bibliography{bibliography}

\end{document}